\documentclass[review]{elsarticle}

\usepackage{url}

\usepackage[lined,boxed,commentsnumbered]{algorithm2e}

\usepackage{booktabs}
\usepackage{hyperref}

\journal{Pattern Recognition}

\begin{document}

\begin{frontmatter}

\title{GPU-Based Computation of 2D Least Median of Squares with Applications to Fast and Robust Line Detection}


\author[mysecondaryaddress]{Gil Shapira\corref{mycorrespondingauthor}}
\cortext[mycorrespondingauthor]{Corresponding author}
\ead{ligaripash@yahoo.com}

\author[mysecondaryaddress,mythirdaddress]{Tal Hassner}
\ead{hassner@openu.ac.il}

\address[mysecondaryaddress]{Department of Mathematics and Computer Science, The Open University of Israel, Israel}
\address[mythirdaddress]{University of Southern California, Information Sciences Institute, CA, USA}

\begin{abstract}
The 2D Least Median of Squares (LMS) is a popular tool in robust regression because of its high breakdown point: up to half of the input data can be contaminated with outliers without affecting the accuracy of the LMS estimator. The complexity of 2D LMS estimation has been shown to be $\Omega(n^2)$ where $n$ is the total number of points. This high theoretical complexity along with the availability of graphics processing units (GPU) motivates the development of a fast, parallel, GPU-based algorithm for LMS computation. We present a CUDA based algorithm for LMS computation and show it to be much faster than the optimal state of the art single threaded CPU algorithm. We begin by describing the proposed method and analyzing its performance. We then demonstrate how it can be used to modify the well-known Hough Transform (HT) in order to efficiently detect image lines in noisy images. Our method is compared with standard HT-based line detection methods and shown to overcome their shortcomings in terms of both efficiency and accuracy.


\end{abstract}

\begin{keyword}
\texttt Robust Regression \sep LMS \sep GPGPU \sep Line Detection \sep Image Processing \sep CUDA \sep Duality
\end{keyword}

\end{frontmatter}


\section{Introduction}

The ability to fit a straight line to a collection of 2D points is key to a wide range of statistical estimation, computer vision and image processing applications. The use of so-called robust estimators to solve this task is of particular interest due to their insensitivity to outlying data points. The basic measure of the robustness of such estimators is their breakdown point: the fraction (up to 50\%) of outlying data points required to corrupt the estimator's accuracy. To date, and despite the many years since its original release, Rousseeuw’s Least Median-of-Squares (LMS) regression (line) estimator~\citep{Rous84} is still among the best known $50\%$ breakdown-point estimators. 

The LMS line estimator (with intercept) is defined formally as follows. Consider a set $S$ of $n$ points $\mathbf{p}_i=(x_i, y_i )_{i=1..n}$ in $R^2$. The problem is to estimate the parameter vector $\mathbf{\theta} = (\theta_1, \theta_2)$ describing the line which best fits the data points by the linear model:
\begin{equation}
y_i = x_i\theta_1 + \theta_2 + e_i,~~~~~i = 1,...,n
\end{equation}
where $(e_1,..., e_n)$ are the (unknown) errors. Given an arbitrary parameter vector $(\theta_1, \theta_2)$, let $r_i = y_i - (x_i\theta_1 + \theta_2)$ denotes the $i$th residual. The LMS estimator is defined to be the parameter vector which minimizes the median of the
squared residuals. This should be contrasted with Ordinary Least Squares (OLS), which minimizes the {\em sum} of the squared residuals.

\begin{figure}[hbt]
\centering
\includegraphics[scale=0.3]{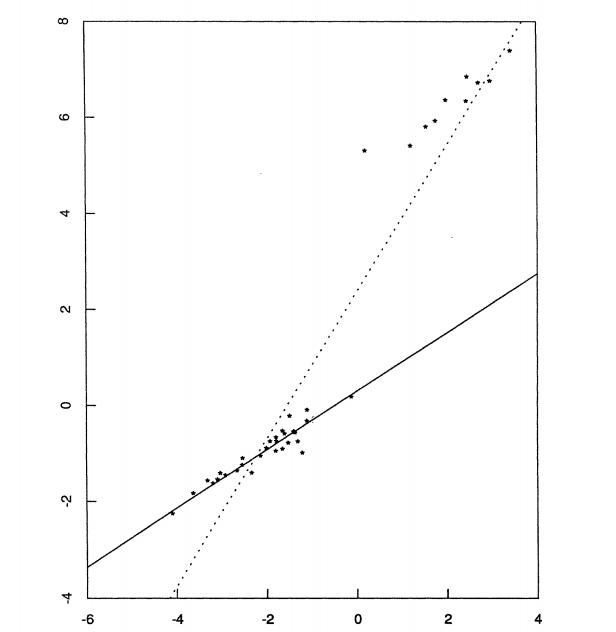}
\caption{{\bf Benefits of high breakdown regression.}
The LMS regression line depicted in dark gray. OLS line depicted in light gray (adapted from~\cite{souvaine1987time}).}
\label{fig:robust_reg}
\end{figure}

Intuitively, if at least 50\% of the points are inliers, the minimized residual median can not be disturbed by higher order statistics residuals and the outliers can be arbitrarily far from the estimate without affecting it. The LMS estimator is also regression, scale and affine equivariant i.e., its estimate transforms ``properly'' under these types of transformations. (See Rousseeuw and Leroy,~\citep{leroy1987robust} for exact definitions.)
The LMS estimator has been used extensively in many applications in a wide range of fields and is considered a standard technique in robust data analysis. 

Our main motivation for studying the LMS estimator stems from its use in computer vision and image processing going as far back as Meer et al.~\cite{meer1991robust}, Netanyahu et al.~\cite{mount2007practical} and Stewart~\cite{stewart1999robust}. Despite the obvious advantages of LMS, it has not been widely adopted, very likely due to its high computational cost. The complexity of LMS, analyzed by Chien and Steiger in~\cite{chien1995some} is shown to have a lower bound of $\Omega­(n log n)$ in the model of algebraic decision trees. In~\cite{gajentaan1995class} Gajentaan and Overmars introduce the concept of 3-sum-hardness: Given $n$ integer numbers, the 3-sum problem is to decide whether three distinct numbers sum up to zero. The best known algorithm for this problem is $O(n^2)$. Every problem that can be reduced to 3-sum is 3-sum-hard. Aiger et al.~\cite{Aiger11} recently showed by that LMS is 3-sum-hard. Thus, a faster than $O(n^2)$ algorithm is unlikely.

Alongside these theoretical results, recent maturing technologies and plummeting hardware prices have made massively parallel, graphics processing units (GPU) standard components in consumer computer systems. Noting this, we propose an exact, fast, CUDA-based, parallel algorithm for 2D LMS estimation. Our method provides a highly efficient and practical alternative to computationally expensive, traditional LMS methods.

\section{Background and related work}
\paragraph{Exact LMS estimation}
Stromberg~\cite{stromberg1993computing} provided one of the early methods for exact LMS with a complexity of $O(n^{d+2}logn)$. More recently, Erickson et al.~\cite{erickson2006least} describe an LMS algorithm with running time of $O(n^dlogn)$. 


Souvaine and Steele~\cite{souvaine1987time} designed two exact algorithm for LMS computation. Both algorithms are based on point-line duality.
One constructs the entire arrangement of the lines dual to the $n$ points and requires $O(n^2)$ time and memory space. The other sweeps the arrangement with a vertical line and require $O(n^2logn)$ time and $O(n)$ space. Edelsbrunner and Souvaine~\cite{edelsbrunner1990computing} have improved these results and give a topological sweep algorithm which computes the LMS in $O(n^2)$ time and $O(n)$ space. As previously mentioned, this result is likely to be optimal. To our knowledge, no GPU-based or other parallel algorithm for computing LMS regression has since been proposed.


\paragraph{LMS approximations}

The high complexity of the exact LMS computation motivated the development of many fast LMS approximation algorithms. The reader is referred to~\cite{erickson2006least,matouvsek1991cutting,mount2000quantile} and~\cite{olson1997approximation} for more detail.


\paragraph{Alternative robust line fitting methods}
Of course, other approaches for fitting lines in the presence of noisy measurements exist. Possibly the most well-known and used methods for line/ hyperplane fitting include the Hough Transform (HT)~\cite{duda1972use} and RANSAC~\cite{fischler1981random}. Both HT and RANSAC are efficient, linear-time complexity algorithms. However neither can be shown to provide a global or even an error-bounded approximate solution to the associated optimization problem.




\subsection{The GPU as a fast parallel processor}
\label{sec:cuda_computation_model}

For over a decade, GPUs have been used for general purpose computation (GPGPU)~\cite{harris2005gpgpu,harris2012gpgpu}. The success of GPU computing lies on both its computation performance superiority over the CPU and the ubiquity and low cost of GPU hardware.
Currently, the theoretical peak badnwidth and gigaflops performance of the best GPU is about seven times higher than the best CPU~\cite{brodtkorb2010state,brodtkorb2013graphics,owens2008gpu}).

This performance gap stems from the architectural differences between the two processors: While CPUs are serial processors, optimized to execute series of operations in order, the GPUs are massively data parallel computing machines: They perform the same computation on many data items in parallel. A substantial performance gain can therefore be achieved if this parallelism can be effectively harnessed.

\paragraph{The CUDA computation model}
Compute Unified Device Architecture (CUDA) is a programming model designed by NVIDIA corporation and implemented by their GPU hardware. The fast LMS method described here is intimately related to CUDA. A comprehensive overview of CUDA, however, falls outside the scope of this paper. We refer the reader to~\cite{sanders2010cuda} for a more detailed treatment of GPGPU computing and CUDA. Below, we provide only a cursory overview of the concepts directly related to our work. 

The GPU computation model is based on a thread hierarchy. Threads are grouped in thread blocks. Threads in the same block can synchronize and cooperate using fast shared memory. Thread blocks are grouped into a grid. The computation kernel is launched on a grid~\cite{brodtkorb2013graphics}. This thread hierarchy is mapped into hardware constructs on the GPU. A thread block runs on a single multiprocessor. Only one thread block can run on a single multiprocessor, and multiple blocks can be executed on a single multiprocessor in a time-slice fashion. The grid of blocks is executed by multiple multiprocessors. Both grid and block can be one- two- or three-dimensional. Each block has a unique identifier within a grid and each thread has a unique identifier within a thread block. These are combined to create a unique global identifier per thread. 


%


One of the main purposes of the massively threaded architecture is to hide memory latencies. GPU memory bandwidth is much higher than CPU memory bandwidth~\cite{brodtkorb2013graphics}, but it still takes a few hundred clock cycles to fetch a single element from the GPU global memory.
This latency is hidden by thread switching: when a thread stalls on a global memory request, the GPU scheduler switches to the next available thread. 

This strategy is most effective when all latencies can be hidden and this requires many threads. The ratio between the total number of thread we create to the maximum number of threads the GPU can handle is referred to as the  GPU {\em occupancy}. The occupancy should be high enough to hide all latencies but once all latencies are hidden, increasing the occupancy may actually adversely affect performance as it also affect other performance metrics.

The GPU executes instructions in 32 way SIMT (single instruction multiple thread) model~\cite{lindholm2008nvidia}: In any given time, the GPU execute a batch of 32 threads on a single multi-processor called a warp. All of the threads in a warp execute the same instruction in a lock step fashion. If the code diverge in a warp (i.e. after a conditional that is true for some of the threads and false for others), the execution is serialized. This can degrade performance severely and should be avoided as much as possible.

\section{Fast CUDA algorithm for 2D LMS computation}
Our algorithm is based on point-line duality and searching line arrangements in the plane. Point-line duality has been shown to be very useful in many efficient algorithms for robust statistical estimation problems (e.g.,~\cite{cole1989optimal,dillencourt1992randomized,edelsbrunner1990computing,erickson2006least} and~\cite{souvaine1987time}). We begin by offering a brief overview of these techniques and their relevance to the method presented in this paper.

\subsection{Geometric interpretation of LMS}

\begin{figure}[hbt]
\centering
\includegraphics[width=0.5\textwidth,natwidth=610,natheight=642]{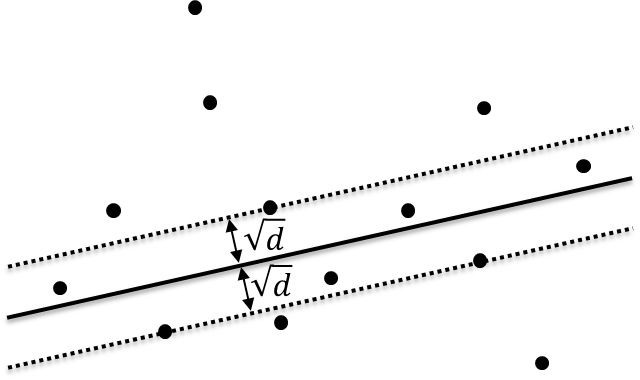}
\caption{LMS geometric interpretation: computing an LMS regression line is equivalent to finding a slab that contains at least half the input points and whose intersection with the y axis is the shortest. $d$ denotes the LMS.}
\label{fig:lms_slab}
\end{figure}

By definition, the LMS estimator minimizes the median of the squared residuals. Geometrically this means that half of the points reside between two lines, parallel to the estimator. Assume $d$ to be the least median of squares, one of the lines is in distance $\sqrt{d}$ above the estimator, and the other is in distance $\sqrt{d}$ below the estimator. 
The area enclosed between these lines is called a slab. Computing the LMS estimator is equivalent to computing a slab of minimum height. The LMS estimator is a line that bisects this slab. See Figure \ref{fig:lms_slab}


\subsection{LMS and point-line duality}

We define the following two planes: Let $x$ and $y$ denote the axes in the {\em primal} plane and $u$ and $v$ denote the axes in the {\em dual} plane. The dual transformation is a mapping between a primal point $p = (a, b)$ to the dual line $v = au -− b$. The dual line is denoted by $p^*$. Conversely, the mapping maps the nonvertical prinal line  $l : y = ax −- b$ to the dual point $(a, b)$. The dual point is denoted by $l^*$. Edelsbrunner~\cite{edelsbrunner1987algorithms} shows that the dual transformation preserves a number of affine properties. In particular, the following properties will be relevant to our later presentation: 



\begin{itemize}
\item \textbf{Order reversing}: Point $p$ lies above/on/below line $l$ in the primal plane if and only if the dual line $p^*$ passes
below/on/above the dual point $l^*$, respectively.

\item \textbf{Intersection preserving:} Lines $l_1$ and $l_2$ intersect at point $p$ in the primal plane if and only if dual line $p^*$ passes through the dual points $l_1^*$ and $l_2^*$.

\item \textbf{Strip enclosure:} A strip of vertical height $h$ bounded by two nonvertical parallel lines $l_1$ and $l_2$ is mapped to a pair
of dual points $l_1*$ and $l_2^*$ such that the segment joining them is vertical and of length h. A point p lies within the strip if and only if the dual line $p^*$ intersects this segment.
\end{itemize}
The first and third properties follow directly from the definition of the dual transformation and the second property follows from the first.


A {\em bracelet} is defined as the vertical segment which results from applying the dual transformation to a strip. Using these properties the LMS can be interpreted in the dual setting: By the strip enclosure property, the LMS dual is equivalent to finding the minimum length bracelet such that $n/2$ of the dual lines intersect it. This is called the LMS bracelet (See~\cite{mount2007practical} for more on these definitions.)



\subsection{Equioscillation and its dual}
\label{Equioscillation}
\begin{figure}[hbt]
\centering
\includegraphics[width=0.8\textwidth,natwidth=610,natheight=642]{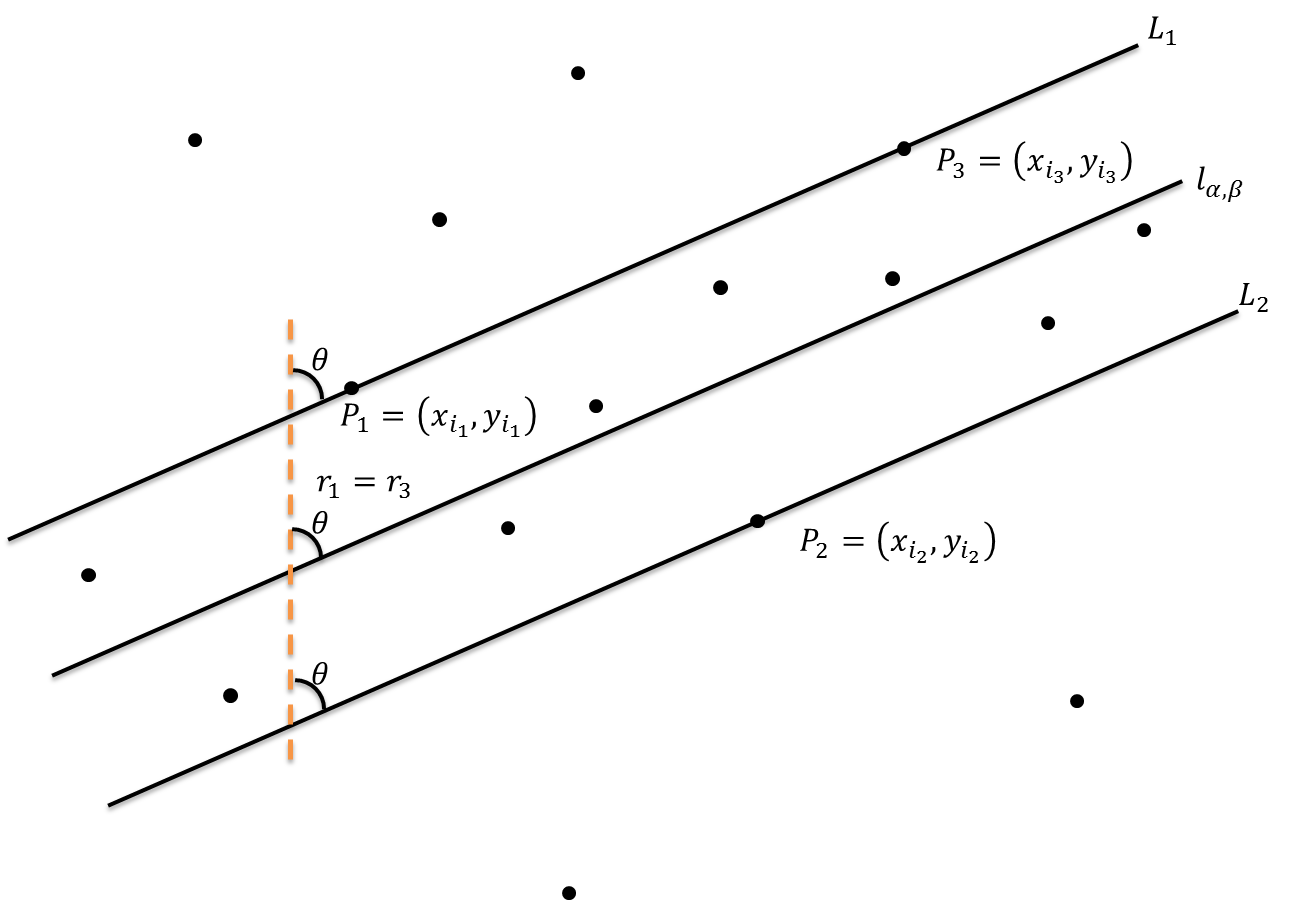}
\caption{Equioscillation. For the given set of data points of size $n = 20$, the LMS regression line $\mathbf{l}_{\alpha, \beta}$ equioscillates relative to $\mathbf{P}_1$, $\mathbf{P}_2$ and $\mathbf{P}_3$. Note that lines $\mathbf{L}_1 \parallel \mathbf{l}_{\alpha, \beta} \parallel \mathbf{L}_2$ are equally spaced (adapted from~\citep{souvaine1987time}).
} 
\label{fig:eqio}
\end{figure}

For any $\alpha$ and $\beta$ the line $\mathbf{l}_{\alpha,\beta} = \{(x, y): y = \alpha x + \beta\}$ defines residuals $r_i$. We say the line $\mathbf{l}_{\alpha,\beta}$ bisects three distinct points $(x_{i_j}, y_{i_j})(j = 1, 2, 3)$ if all of the $r_{i_j}$ are of the same magnitude $r$ but not all have the same sign. If $x_{i_1} < x_{i_2} < x_{i_3}$ and $r_{i_1} = -r_{i_2} = r_{i_3}$ we say $l_{\alpha,\beta}$ {\em equioscillates} with respect to the points. These concepts are illustrated in Figure~\ref{fig:eqio}.

It was proved in Steele and Steiger~\cite{steele1986algorithms} that LMS regression line must be an equioscillating line relative to some triple of data points. On the dual setting, this means that the LMS bracelet segment has one end on the intersection of two lines. 
 
Our algorithmic problem in the dual can now be expressed as follows: Given $n$ lines $\mathbf{L}_i (1 \leq i \leq n)$ in general position in the plane with intersection points $\mathbf{P}_{i_j} (1 \leq i<j \leq n)$, find the line $\mathbf{L}^*$ and intersection point $\mathbf{p}^*$ such that, among all line-point pairs $(\mathbf{L}, \mathbf{P})$ that have exactly $K$ of the $\mathbf{L}_i$ cutting the vertical segment $S$ joining $\mathbf{L}$ and $\mathbf{P}$, the pair $(\mathbf{L}^*, \mathbf{P}^*)$ has the smallest vertical distance. In other words, the minimal bracelet we are searching for has an intersection points on it's upper or lower end.

\subsection{Parallel algorithm for the LMS}
Our parallel algorithm is based on the one proposed in~\cite{edelsbrunner1990computing}. Using the equioscillation property stated in Section~\ref{Equioscillation} it computes all the bracelets hanging on an intersection point and then concurrently searches for the minimum bracelet among them. Pseudo code describing this process is given in Algorithm~\ref{algo_lms} below.


\IncMargin{1em}
\begin{algorithm}
\LinesNumbered
\SetKwFor{ParForEach}{foreach}{parallel do}{endfch}
\SetKwFor{ParFor}{for}{parallel do}{endfor}
\SetKwData{Left}{left}\SetKwData{This}{this}\SetKwData{Up}{up}
\SetKwFunction{Union}{Union}\SetKwFunction{FindCompress}{FindCompress}
\SetKwInOut{Input}{input}\SetKwInOut{Output}{output}
\SetKwInOut{ParallelDo}{Parallel do}
\Input{set $P$ of points, $P \subset \Re_2$}
\Output{The LMS regression line and slab height}
\BlankLine

\ParForEach{point $p_i \in P$ }
{compute it's dual $l_i$ and insert to $L$}

\ParForEach{pair of lines $l_i, l_j \in L ( i \neq j )$}
{
compute the intersection point $ip_{i,j}$

$I \leftarrow ip$
}

\ParForEach{intersection point $ip \in I$}
{
\ParForEach{$l \in L$}
{
compute the intersection point $x$ of $l$ with the vertical line that pass through $ip$.

$X \leftarrow x$.
}
Parallel sort the points in $X$ by their $y$ coordinate.

Assume point $ip$ has order $k$ in the sorted sequence.

$BraceletSecondPoint \leftarrow X[(k + \frac{\left\vert P \right\vert}{2}) \bmod { \left\vert P \right\vert}]$

save the founded bracelet data for the current intersection point $ip$ in an array $BraceletArray$
}

Use parallel reduction to find minimum length bracelet on $BraceletArray$

Translate minimal bracelet data back to the primal plane and return LMS regression line equation and slab height.

\caption{Pseudo code for our proposed parallel LMS method.}\label{algo_lms}
\end{algorithm}


\subsubsection{Algorithm Correctness}
By the equioscillation property of the LMS regression line (Section~\ref{Equioscillation}), it is assured that the minimum bracelet has one end on the intersection of two lines. The algorithm exhaustively searches for all bracelets that have this property, thus the global minimum bracelet must be found and the algorithm is correct.

\subsection{Algorithm performance}\label{alg_perf}
Assuming the PRAM model~\cite{karp1988survey}, and a parallel machine with unbounded number of processors, parallel computation of all the points duals (lines 1-3) takes $O(1)$ time (each processor computes one point dual). Computing the intersection points of all line pairs (lines 4-7) also requires $O(1)$ time (each processor computes the intersection point of one pair of lines). For each intersection point we find the intersection of all the lines with the a vertical line that pass through the intersection point (lines 9-12). This is also a $O(1)$ time task. A bitonic parallel sorter (line 13) takes $O({log}^2n)$ (see Batcher~\cite{batcher1968sorting}). On the final step, a parallel reduction (line 18) is performed on an array in the size of the number of intersection points, which is $O(n^2)$. This parallel reduction takes $O(log n)$ time. The overall time complexity therefore equals to the time of the bitonic sorter: $T_\infty = O({log}^2n)$.
This should be compared to the optimal sequential algorithm, which requires $O(n^2)$ time.

The total work required by the algorithm is sorting $n$ elements for each of the $O(n^2)$ intersection points which amounts to $O(n^3 log n)$ computations. This is also equals to the time complexity $T_1$ of the algorithm running on one processor. On a machine with a bounded number of processor the time complexity is asymptotically equal to $T_1$ hence for large input size, the parallel version is expected to do worse than the optimal $O(n^2)$ sequential algorithm.

Finally, the space complexity of our method is $O(n^2)$, as we must save all intersection points in an array.

\subsection{Details of the CUDA implementation}
In accordance with Section~\ref{sec:cuda_computation_model}, to achieve optimal performance using the CUDA platform we set the following design goals:

\begin{itemize}
  \item Maximize occupancy.
  \item Memory coalescence - adjacent threads calls adjacent memory locations.
  \item Bank conflict free shared memory access pattern.
\end{itemize}

\begin{figure}[hbt]

\centering
\includegraphics[width=0.8\textwidth,natwidth=610,natheight=642]{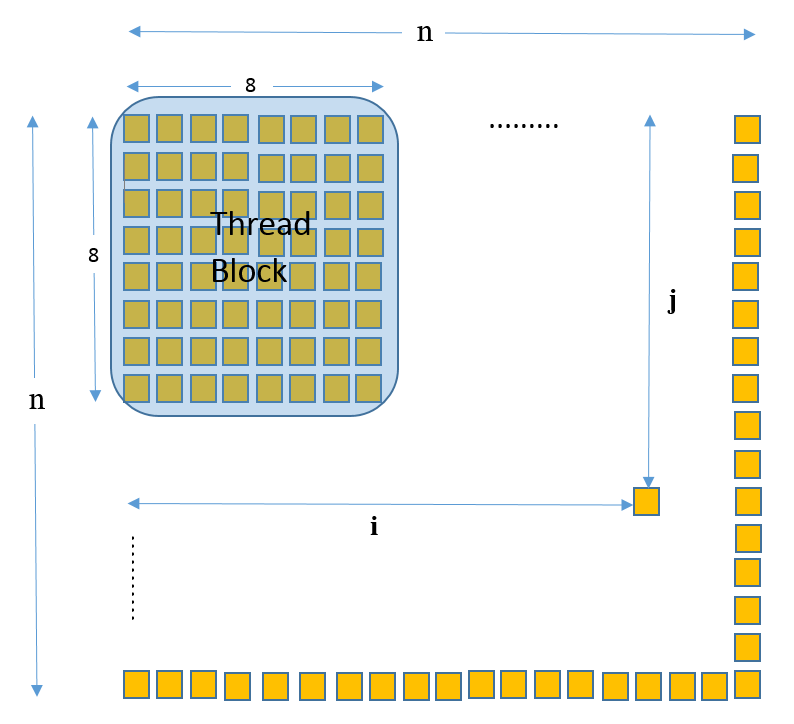}
\caption{The thread grid for computing line intersections. Each thread block has $8\times 8$ threads. In the grid there are $\frac{n}{8}\times \frac{n}{8}$ thread blocks. Thread $t_{i,j}$ calculates the intersection of line $i$ with line $j$. $n = \left\vert\mathbf{L}\right\vert$}\label{fig:threadgrid}
\end{figure}

The first part of the algorithm requires calculating the intersection point of each pair of lines. To do that in CUDA we create the following thread hierarchy: Each thread block contains $8\times 8$ threads and each grid contains $\frac{\left\vert L \right\vert}{8}\times \frac{\left\vert L \right\vert}{8}$ thread block. This creates the thread structure displayed in Figure~\ref{fig:threadgrid}.

Here, memory access is coalesced as thread $t_{i,j}$ reads line $\mathbf{l}_i$ and $\mathbf{l}_j$ while the next thread in the row $t_{i,j+1}$ reads $\mathbf{l}_i$ and $\mathbf{l}_{j+1}$ which resides right next to $\mathbf{l}_j$ in the line array. To save redundant access to global memory, threads in the first row and first column of each thread block read the respected lines from global memory to shared memory of the block. This saves $2\times(64 - 2\times 8) = 96$ global memory access per block. To avoid computing line intersections twice, only threads $t_{i,j}$ where $i > j $ perform the computation while the rest do nothing. Each thread computes the intersection of 2 lines in $\mathbf{L}$ with the vertical line passing through it's intersection point. The intersection points are stored in shared memory. 

\begin{figure}[hbt]
\centering
\includegraphics[width=0.8\textwidth,natwidth=610,natheight=642]{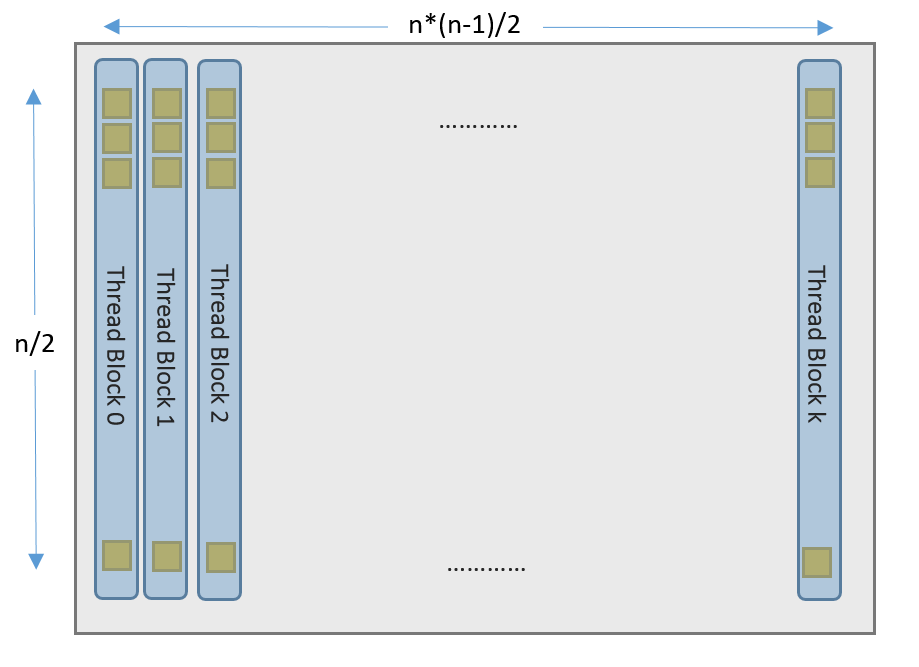}
\caption{The thread structure for bracelet calculation per intersection point. Each thread block computes the bracelet for its designated intersection point. The thread grid contains $\frac{n\times (n-1)}{2}$ thread blocks. Each thread block contains $n$ threads.}
\label{fig:threadstructure}
\end{figure}

In the second part of the algorithm we compute the bracelet that hangs on every intersection point found earlier (see Algorithm~\ref{algo_lms}, lines 8-17). To do that, we assign a thread block to compute each bracelet (see Figure \ref{fig:threadstructure}). The thread block dimension is $(1, \frac{\left\vert L \right\vert}{2})$. Each thread computes the intersection points of two lines from $\mathbf{L}$ with the vertical line which passes through the intersection point assigned to the thread block. later, the thread block perform fast parallel bitonic sort (see~\cite{peters2010fast}). Parallel bitonic sort operates only on power of two data items and so we allow only power of two input size. When the input size does not meet this requirement, this limitation can be easily dealt with by padding: The thread grid has dimension $(k,1)$ where $k$ is the number of point intersections ($k = \frac{n \times (n - 1)}{2}$). 

\section{Results}
Our implementation of the CUDA LMS method, described in Algorithm~\ref{algo_lms} is publicly available online\footnote{Available:~\url{https://github.com/ligaripash/CudaLMS2D.git}}. We have compared it with the state of the art, topological sweep algorithm of Eddelsbrunner and Souviene~\cite{edelsbrunner1990computing}. A software implementation of a variant of this algorithm is given by Rafalin~\cite{rafalin2002topological}. In our comparison, both algorithms where provided with the same random point set of sizes 128, 256 and 512 points. Correctness of our algorithm is verified by comparing its output for each point set with the output of the topological sweep method. 

\begin{table}[ht]
\begin{center}
\caption{Performance comparison between the state of the art, serial topological sweep algorithm of~\cite{edelsbrunner1990computing} ({\em Top Sweep}) and our own CUDA LMS implementation.}
\label{CUDAperfTable}
\begin{tabular}{cccc}
\toprule
\textbf{\begin{tabular}[c]{@{}c@{}}Input Point \\ Set Size\end{tabular}} & \multicolumn{1}{l}{\textbf{\begin{tabular}[c]{@{}l@{}}Top Sweep \\ Time {[}ms{]}\end{tabular}}} & \multicolumn{1}{l}{\textbf{\begin{tabular}[c]{@{}l@{}}CUDA LMS\\ Time {[}ms{]}\end{tabular}}} & \multicolumn{1}{l}{\textbf{Speedup Factor}} \\ \hline
128 & 42.52 & 0.98 & 46.44                                        \\ 
256 & 172.12 & 4.22 & 40.78                                        \\ 
512 & 691.35 & 35.3 & 19.57                                        \\ \bottomrule
\end{tabular}
\end{center}
\end{table}

Results are reported in Table~\ref{CUDAperfTable}. Evidently, in all cases the speedup factor of our proposed method was substantial. All measurements were taken using a machine with CPU type: intel E6550 2.3 Ghz and GPU type: NVIDIA GTX 720. These results enable robust estimation in real time image processing applications. In the following section we introduce a method for fast and robust line detection in noisy images, based on our CUDA LMS algorithm. We note that the speedup factor appears to drop with set size. This is due to the fact that the sequential algorithm is asymptotically faster, as we discuss in Section~\ref{alg_perf}.

\section{CUDA LMS for fast and robust line detection in images}


Line detection in image data is a basic task in pattern recognition and computer vision used for both data reduction and pre-processing before higher level visual inference stages. Most existing line detection methods are variants of the Standard Hough Transform (SHT). SHT is a one-to-many method where each feature in the image votes for all lines passing through it. Theses votes are accumulated in a $nD$ array called an {\em accumulator}. Later, peeks in the accumulator are detected and the lines they represent are returned. The computational complexity of SHT is
$O(N\delta^{d-1})$, where N is the number of given points and $\delta$ is the (discrete) dimensions of the image. This is linear for a fixed $\delta$ and a fixed $d$. The reader is referred to~\cite{mukhopadhyay2014survey} for the latest survey on HT and its variants.


Despite its popularity, SHT has some well known shortcomings (see~\cite{mukhopadhyay2014survey}):
\begin{itemize}
\item \textit {Accuracy/Performance trade off} - For accurate line detection, a high resolution accumulator is needed. As a result the voting procedure is slow and time performance suffers. A low resolution accumulator, on the other hand, means faster computation but with diminished accuracy. To address the time performance problem several randomization variants have been suggested. Two such popular algorithms are Probabilistic HT (PHT)~\cite{kiryati1991probabilistic} and Randomized HT (RHT)~\cite{xu1990new}. These methods trade a small amount of detection accuracy for improved computation speed.

\item \textit {Accumulator resolution/Peak Selection trade off} - There are various sources of noise in an image. Sensor noise, optical distortions and image quantization all result in a spread of votes around the actual bin in the parameter space, which, in turn, leads to inaccuracy in peak detection. The higher the accumulator resolution the more accute this peak spreading phenomena becomes and it is more difficult to locate the best line in the cluster~\cite{shapiro2006accuracy}. When accumulator resolution is low, peak spreading is lower and finding the best line in the cluster is easier but this comes at the cost of diminished accuracy. To deal with the peak spreading problem several method have been described, proposing to analyze patterns in the accumulator array in order to obtain enhance accuracy~\cite{atiquzzaman1994complete,furukawa2003accurate}, and~\cite{ji2011novel}.


\end{itemize}

We propose to address both issues by following up the process performed by the SHT, with our CUDA based LMS method, applied separately to the point set of each putative line. 

Specifically, we begin by applying the SHT using the Canny edge detector to extract image feature points and a polar coordinate line voting accumulator array~\cite{duda1972use}. A coarse voting grid is used as it allows for fast computation. As mentioned, this typically implies a compromise in the accuracy of the detected line. Here, however, we follow line detection based on accumulator votes with an LMS step, applied separately to the set of points supporting each line. 

The rationale here is simple: coarse accumulator array coordinates imply poor line localization and introduction of noisy measurements. By assuming an outlier ratio of less than 50$\%$ of the points contributing to each peek, we can apply LMS in order to ignore outlying points and recover accurate line parameter estimates. 

The benefits of this approach can be summarized as follows: 

\begin{itemize}
\item Accuracy and robustness - up to 50$\%$ of the feature points voting for a specific cell in the Hough accumulator can be attributed to noise or another line structure and still the correct line can be accurately recovered.

\item Speed - Because we use the Hough accumulator to locate feature points and not for direct computation of line equation as in SHT, we can use fast, low resolution accumulator. The basic building block of this method, the CUDA LMS is fast (Table~\ref{CUDAperfTable}) and can be further accelerated by using random sampling to sample feature points voting for a specific Hough cell.

\end{itemize}

This method can be incorporated with other HT variants which uses the accumulator peak cell to recover the feature points in the image that voted for the cell and recover the line by means of Least Square method or edge following such as in progressive probabilistic Hough Transform (PPHT)~\cite{galambos2001gradient}. This can be done by replacing regular least square used by these methods with LMS.

\subsection{Line detection results}

\begin{figure}[hbt]
\centering
\fbox{\includegraphics[width=0.5\linewidth]{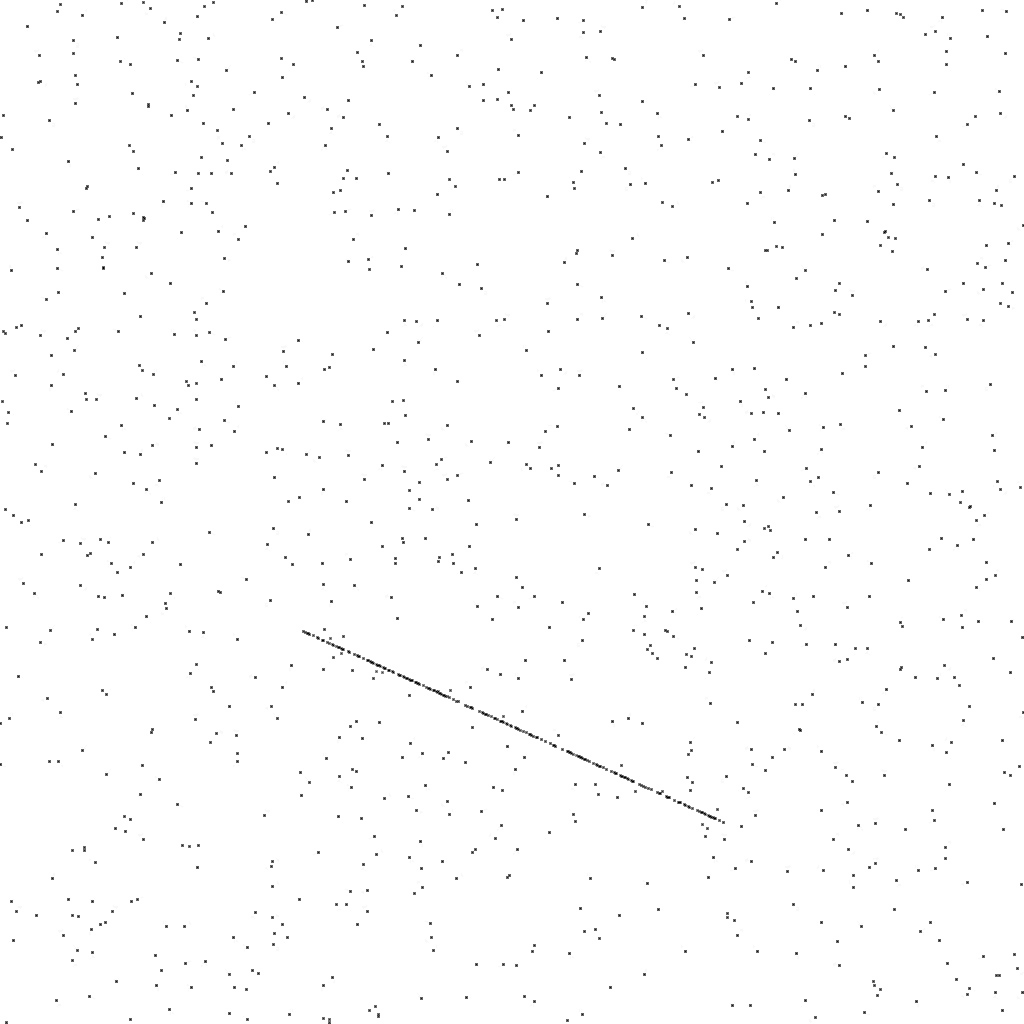}}
\caption{A synthetic image example from the ones used in our line detection experiments. Here, random noise was applied with probability = 0.001 and line sampling probability = 0.5}
\label{fig:input_image}
\end{figure}

We compare our modified HT with the following well used, existing alternatives:
\begin{itemize}
\item Regular SHT
\item A method similar to the one proposed here, but using Least Squares instead of LMS to calculate the regression line for the image points that voted for the peak accumulator bin (OLS method). 
\item Our CUDA LMS based method
\end{itemize}
Our tests were performed using identical data: the same edge detector, voting process and accumulator peek detection process were used for all methods tested. To obtain accurate ground truth data we used synthetic $1024\times 1024$ images. These contained random noise in various probabilities and one line segment with points sampled on the line with probability 0.5 (See Figure~\ref{fig:input_image}).

\begin{figure}[hbt]
\centering
\includegraphics[width=\linewidth]{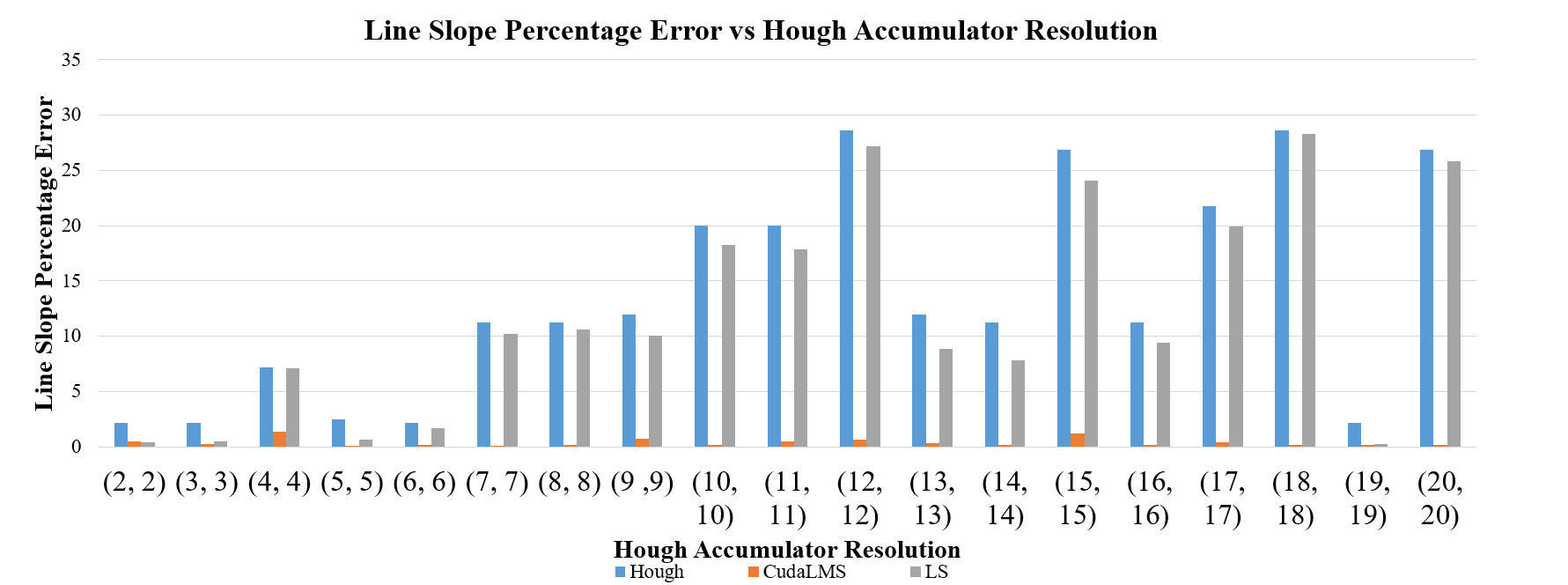}
\caption{Comparison of line slope error for different Hough accumulator resolutions and line estimation methods. The CudaLMS error is uncorrelated with Hough resolution}
\label{fig:line_slope_error_vs_resolution}
\end{figure}

Figure~\ref{fig:line_slope_error_vs_resolution} reports the percentages of error in slope and intercept for the three methods, for a single random line in various HT accumulator resolutions. Our CUDA LMS regression method has an average slope error of $0.37\%$, which two orders of magnitude smaller than the one reported by using standard Least Squares ($12.4\%$). 

Also noteworthy is that there seems to be little or no correlation between the resolution of the accumulator bins and accuracy. This can be explained by noting that as bin sizes increase, the number of supporting points in each bin also increases. So long as outlier points in each bin do not account for more than $50\%$ of its members, the LMS estimate will remain accurate. 

\begin{figure}[hbt]
\centering
\includegraphics[width=\linewidth]{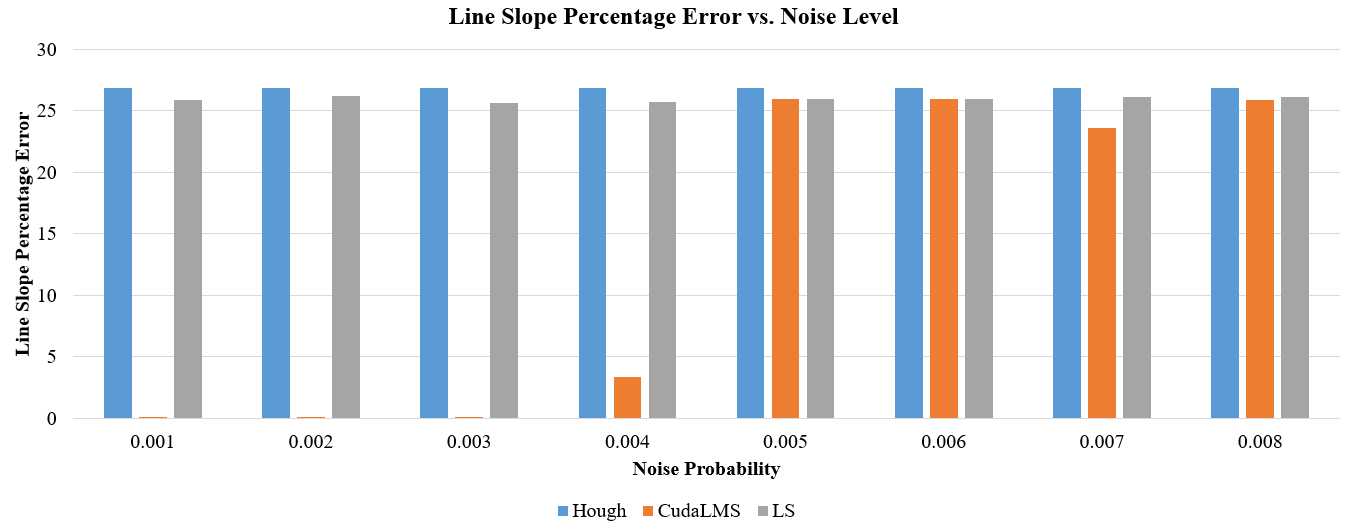}
\caption{Line slope percentage error vs. noise level. On noise levels lower than 0.004, noise points account for less than $50\%$ of the supporting points hence our CUDA LMS based method is very accurate. Above noise level 0.004 this situation is reversed with noise majority which brakes the LMS estimate.}
\label{fig:error_vs_noise}
\end{figure}

Next, we measure the influence of noisy line measurements by adding increasing amounts of noise to the points sampled along a line in the image. Results are reported for the three methods in Figure~\ref{fig:error_vs_noise}.

Evident from Figure~\ref{fig:error_vs_noise} is that LMS line estimates are accurate, with less than $0.14\%$ error in the slope estimate.
Also noteworthy is that for a noise level of 0.004 the error is not negligible with $3.38\%$. At this point more than half of the supporting points are attributed to noise rather than the line itself, and the estimate falters.

\begin{figure}[hbt]
\centering
\fbox{\includegraphics[width=\linewidth]{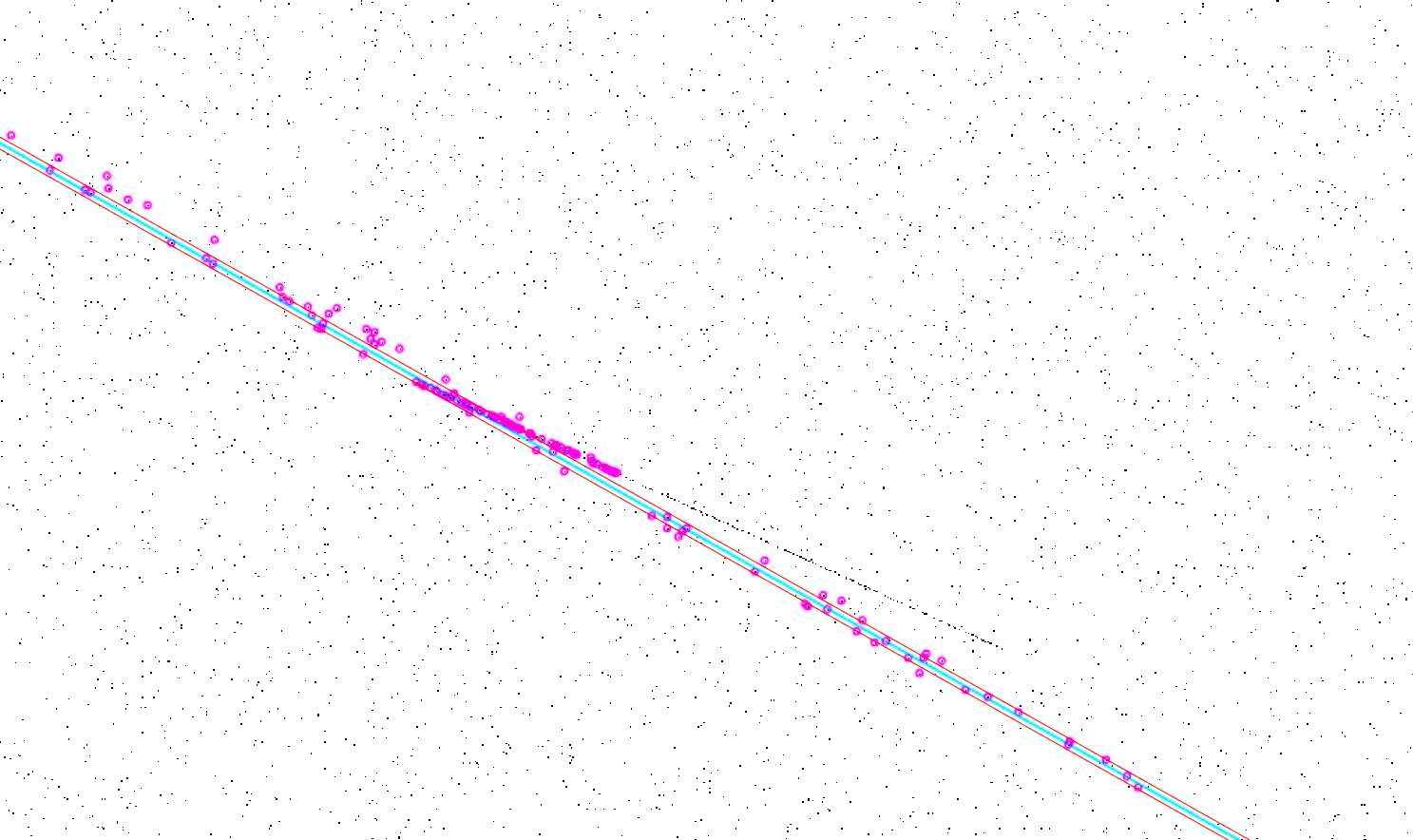}}
\caption{Synthetic image with random noise probability = 0.006 and line sampling probability = 0.5. $\delta\rho$ = 20, $\delta\theta$ = 20. The supporting features are marked in purple circles. LMS slab is marked in red and the LMS line in cyan. More than half of the supporting features are attributed to noise and the LMS estimate breaks down.}
\label{fig:lms_error}
\end{figure}

\begin{figure}[hbt]
\centering
\fbox{\includegraphics[width=\linewidth]{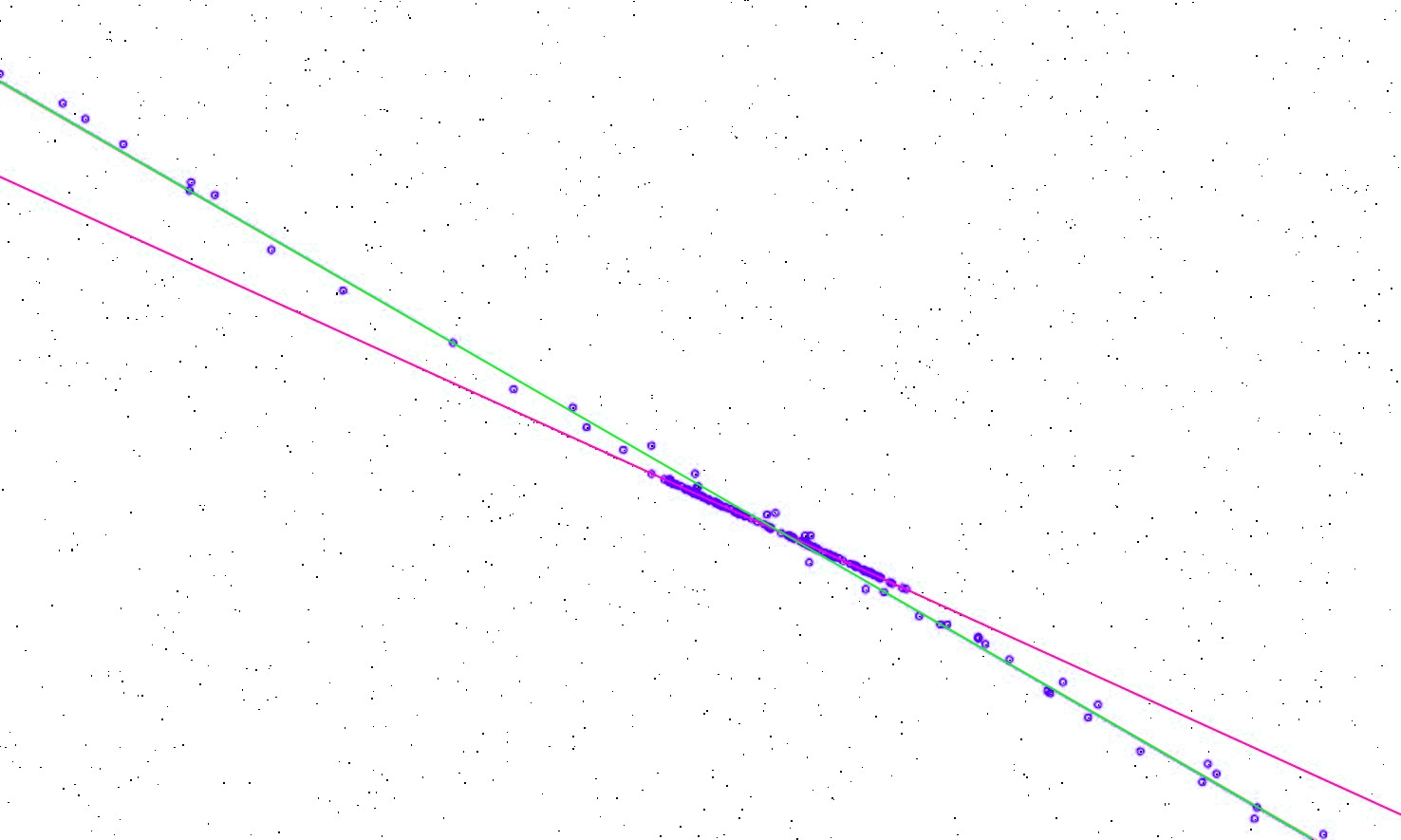}}
\caption{Synthetic image with random noise probability = 0.002 and line sampling probability = 0.5. $\delta\rho$ = 20, $\delta\theta$ = 20. The supporting features are marked in purple circles. LMS line is marked in red. SHT line is marked in green. Less than half of the supporting features are attributed to noise and so our CUDA LMS regression estimation is accurate.}
\label{fig:lms_good_estimation_noise_0002}
\end{figure}

We illustrate a failure of the LMS estimator for a noise ratio of $0.006\%$ in Figure~\ref{fig:lms_error}. A correct LMS estimate with $0.002\%$ noise is provided in Figure~\ref{fig:lms_good_estimation_noise_0002}.

\begin{figure}[hbt]
\centering
\includegraphics[width=.7\linewidth]{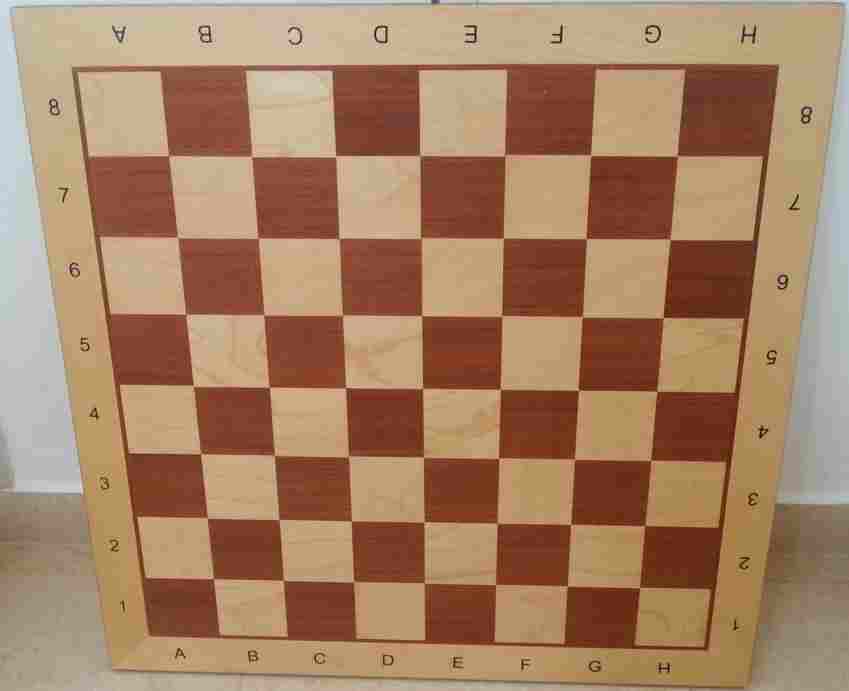}
\caption{Input chess board image.}
\label{fig:input_chessboard}
\end{figure}

Finally, we test performance on a real photo (Figure~\ref{fig:input_chessboard}) with increasing amounts of synthetically added noise. The Hough resolution used is $\delta\rho = 5, \delta\theta = 2$. The ground truth lines were marked by hand.

\begin{figure}[hbt]
\centering
\includegraphics[width=\linewidth]{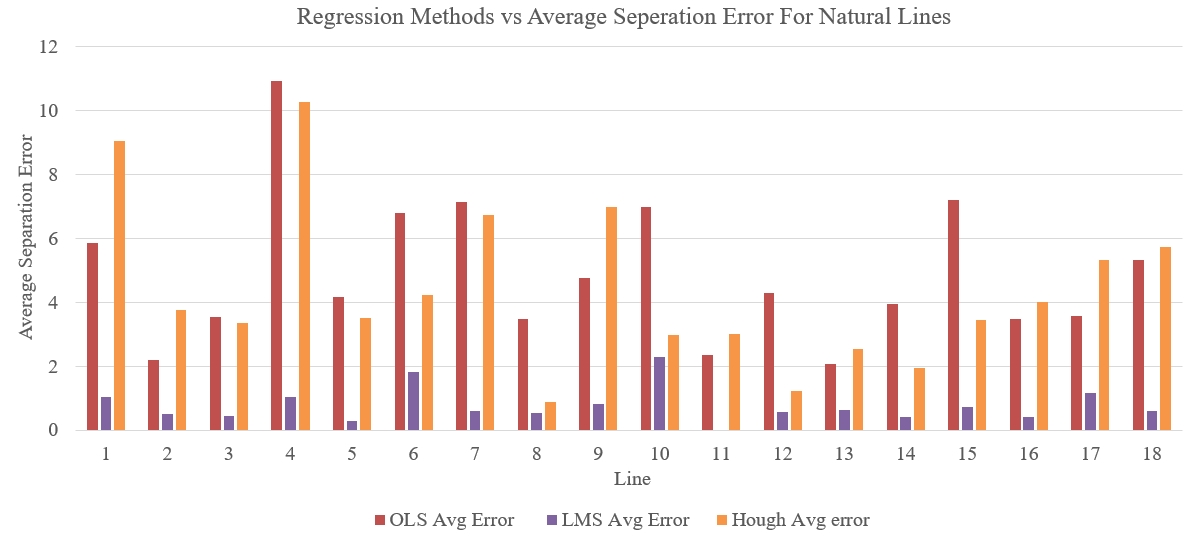}
\caption{The error of the regression lines relative to the ground truth for each line in the image. Our CUDA LMS based method is the most accurate.}
\label{fig:natural_image_results}
\end{figure}

Our results are reported in Figure~\ref{fig:natural_image_results}. For each line estimate, the error was measured by computing the average vertical separation between the estimate and the ground truth (for horizontal lines), and the average horizontal separation between the estimate and the ground truth for vertical lines. Table~\ref{table:natural_summary} additionally summarize the average and standard deviation of the error for the three methods.

\begin{table}[ht]
\begin{center}
\caption{Summary of accuracy comparing different methods on the real photo experiments.}
\label{table:natural_summary}
\begin{tabular}{lcc}
\toprule
        & \multicolumn{1}{l}{Average Error [Pixels]} & \multicolumn{1}{l}{Error STDV [Pixels]} \\ \hline
Hough   & 4.393                              & 2.658                           \\ 
OLS     & 4.893                              & 2.840                           \\ 
CUDA LMS & 0.765                              & 0.404                           \\ \bottomrule
\end{tabular}
\end{center}
\end{table}

\begin{figure}[hbt]
\centering
\includegraphics[width=.7\linewidth]{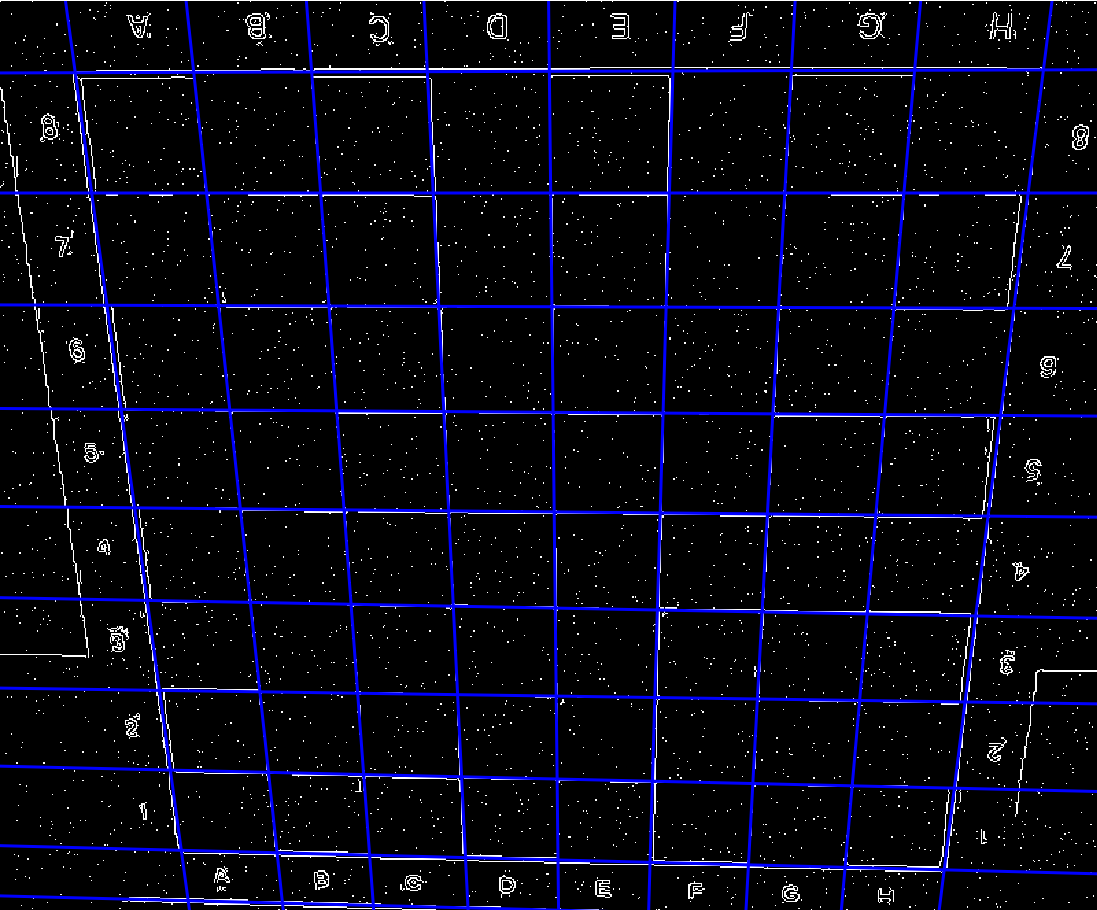}
\caption{Example of lines detected using our CUDE LMS based method.}
\label{fig:natural_lines}
\end{figure}

We can observe That our CUDE LMS based line detection method is extremely accurate, with average error of 0.765 pixels and standard deviation 0.404 pixels. Example lines detected by our method are provided in Figure~\ref{fig:natural_lines}.

\begin{figure}[hbt]
\centering
\includegraphics[width=.7\linewidth]{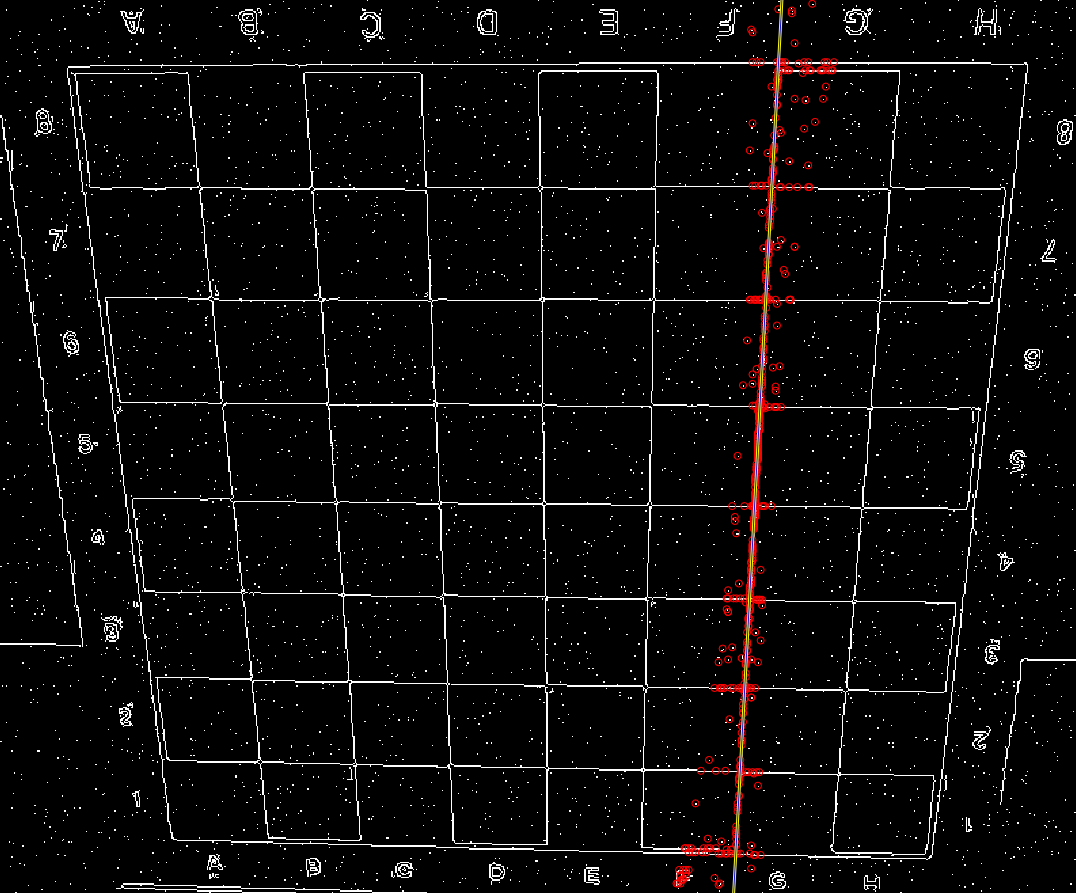}
\caption{The edge image with the supporting features of one of the lines, and our CUDA LMS regression line. Although the supporting features are spread considerably, the line estimate remains accurate as the majority of the features reside on the true line.}
\label{fig:natural_supporting_features}
\end{figure}

Figure~\ref{fig:natural_supporting_features} further illustrates the input edge image with supporting features for one of the lines and the LMS line and bracelet. We can observe that although the supporting features are widely spread around the line, the LMS estimate remains highly accurate.

\section{Conclusions}

This paper addresses one of the fundamental tools of image processing: line detection. Despite being a provably robust means of estimating lines from noisy measurements, the Least Median of Squares method has largely been ignored due to its high computational cost, in favor of faster approximate solutions.

Motivated by the widespread use of GPU hardware in modern computer systems, we propose a CUDA, GPU based implementation of the LMS method. We then show it to be both extremely fast and more accurate than its widely used alternatives. We further show how our CUDA base LMS method may be combined with an existing line detection method, the Hough Transform, in order to provide far more accurate line estimates at very small computational costs. Our method can be used as a substitute for existing, approximate line detection techniques. In order to promote its adoption by others as well as reproduction of our results, we share our code here: \url{https://github.com/ligaripash/CudaLMS2D.git}



\bibliography{fast_gpu_based_2d_lms_and_applications}

\end{document}